\documentclass[conference]{IEEEtran}
\IEEEoverridecommandlockouts

\usepackage[T1]{fontenc}
\usepackage[utf8]{inputenc}        
\usepackage{newtxtext,newtxmath}   

\input{glyphtounicode}
\usepackage{cite}
\usepackage{amsmath}               
\usepackage{algorithmic}
\usepackage{graphicx}
\usepackage{textcomp}
\usepackage{xcolor}
\usepackage{tabularx}
\usepackage{booktabs}
\usepackage{multirow}
\usepackage{url}
\usepackage{tikz}
\usetikzlibrary{shapes,arrows,positioning}

\usepackage[hidelinks]{hyperref}

\def\BibTeX{{\rm B\kern-.05em{\sc i\kern-.025em b}\kern-.08em
    T\kern-.1667em\lower.7ex\hbox{E}\kern-.125em X}}

\begin{document}

\title{Beyond Appearance: Transformer-based Person Identification from Conversational Dynamics}

\author{
    \IEEEauthorblockN{Masoumeh Chapariniya\IEEEauthorrefmark{1}, 
    Teodora Vuković, Sarah Ebling, Volker Dellwo}
    \IEEEauthorblockA{\IEEEauthorrefmark{1}Department of Computational Linguistics, University of Zurich, Zurich, Switzerland \\
    masoumeh.chapariniya@uzh.ch, teodora.vukovic2@uzh.ch, \\
    ebling@cl.uzh.ch, volker.dellwo@uzh.ch}
}

\maketitle

\begin{abstract}

This paper investigates the performance of transformer-based architectures for person identification in natural, face‑to‑face conversation scenario. We implement and evaluate a two-stream framework that separately models spatial configurations and temporal motion patterns of 133 COCO WholeBody keypoints, extracted from a subset of the CANDOR conversational corpus. Our experiments compare pre-trained and from-scratch training, investigate the use of velocity features, and introduce a multi-scale temporal transformer for hierarchical motion modeling. Results demonstrate that domain-specific training significantly outperforms transfer learning, and that spatial configurations carry more discriminative information than temporal dynamics. The spatial transformer achieves 95.74\% accuracy, while the multi-scale temporal transformer achieves 93.90\%. Feature‑level fusion pushes performance to 98.03\%, confirming that postural and dynamic information are complementary. These findings highlight the potential of transformer architectures for person identification in natural interactions and provide insights for future multimodal and cross-cultural studies.
\end{abstract}

\section{Introduction}
Humans reveal identity not only through static appearance but also through the rhythms of everyday conversation—subtle head turns, hand beats, and micro‑expressions that arise from long‑term motor habits and are hard to imitate.  Traditional biometrics focus on \emph{static} traits (facial geometry, fingerprints, hand shape) that suffer under pose changes, lighting, or synthetic manipulation~\cite{baisa}.  To overcome these limits, research has shifted toward \emph{dynamic} cues such as gait~\cite{gait},
voice~\cite{voice}, and signature motion~\cite{Alrawili,Battisti}.  
Hybrid systems that fuse static and dynamic evidence already show clear gains—for instance, combining body measurements with gait improves recognition over either cue alone~\cite{L. Wang}.  Building on this trend, we investigate natural conversational gestures as a rich, yet under‑explored, dynamic signal for person identification.

Recent research has demonstrated the significant potential of dynamic facial features for person identification, complementing traditional static approaches. Farhadipour et al. \cite{Farhadipour} explored multimodal person identification by integrating facial and voice features through sensor-level, feature-level, and score-level fusion strategies, employing gammatonegram representations with x-vectors for voice and VGGFace2 for facial modality.

Hill and Johnston \cite{Hill} established that head and facial movements convey identity-related information beyond static features, demonstrating that rigid head movements are particularly effective for individual distinction while non-rigid facial movements aid in sex categorization. Building on this foundation, Girges et al. \cite{Girges} used marker-less motion capture to investigate facial motion's role in identity recognition, showing that participants could accurately recognize identities using motion patterns alone, even when appearance cues were eliminated.

Further research by \cite{Dobs} examined different facial movement types, revealing that conversational facial movements transmitted the most identity-related information compared to purely emotional expressions, suggesting their utility for biometric systems in dynamic environments. These foundational studies \cite{Hill, Girges, Dobs} primarily relied on motion capture techniques and human experiments rather than computational approaches.

Recent computational advances have leveraged deep learning for dynamic face identification. Papadopoulos et al. \cite{Papadopoulos} proposed a spatio-temporal graph convolutional network (ST-GCN) framework using 3D facial landmarks on the BU4DFE dataset \cite{X. Zhang}, while Kay et al. \cite{Kay} demonstrated that facial micro-expressions provide unique identity cues using SlowFast CNN models on CASME II \cite{CASME} and SAMM \cite{Samm} datasets. Saracbasi et al. \cite{Saracbasi} introduced the MYFED database to analyze six basic emotions for person identification, emphasizing dynamic emotional expressions' role in improving biometric reliability.

Despite these advances, several critical limitations persist in current person identification research. Most existing studies rely on controlled laboratory settings where participants are explicitly instructed to express specific emotions or perform predetermined gestures \cite{rack2022comparison}. These datasets, while providing high-quality recordings under ideal conditions with consistent lighting and camera angles, fail to capture the natural variability and spontaneous dynamics present in real-world interactions. Consequently, there exists a significant domain gap between laboratory-based models and practical applications in unconstrained environments.

Furthermore, previous approaches have predominantly employed traditional deep learning architectures or relied on human perceptual studies, with limited exploration of advanced sequence modeling techniques such as transformer architectures. While recent work like Face-GCN has applied graph convolutional networks to dynamic 3D face identification, the potential of transformer models—which have demonstrated remarkable success in sequence modeling tasks—remains largely unexplored for person identification using natural conversational dynamics. Additionally, existing methods typically focus on isolated facial features rather than leveraging the rich information available from holistic upper body movements and their temporal relationships during conversational interactions.

To address these limitations and bridge the gap between controlled laboratory studies and real-world applications, this paper proposes a novel transformer-based approach for person identification using natural conversational gestures and dynamic facial expressions. Our contributions are as follows:

\begin{itemize}
    \item We introduce spatial and temporal transformer architectures specifically designed to capture discriminative patterns from natural conversational dynamics, eliminating the need for posed emotional expressions or controlled gestures.
    
    \item We propose a multi-scale temporal transformer that learns hierarchical motion dynamics by processing gestures at multiple temporal resolutions, effectively capturing both fine-grained micro-expressions and extended gestural sequences.
    
    \item Investigation of different transformer architectures for capturing identity cues from conversational movements, demonstrating that the fusion of temporal and spatial transformers achieves optimal performance.
    
    \item To our knowledge, this represents the first systematic investigation of transformer-based architectures for person identification using holistic upper body and facial dynamics in natural conversational settings.
\end{itemize}
\section{Methodology}

\subsection{System Architecture Overview}

\begin{figure*}[h!]
    \centering
    \includegraphics[width=\textwidth]{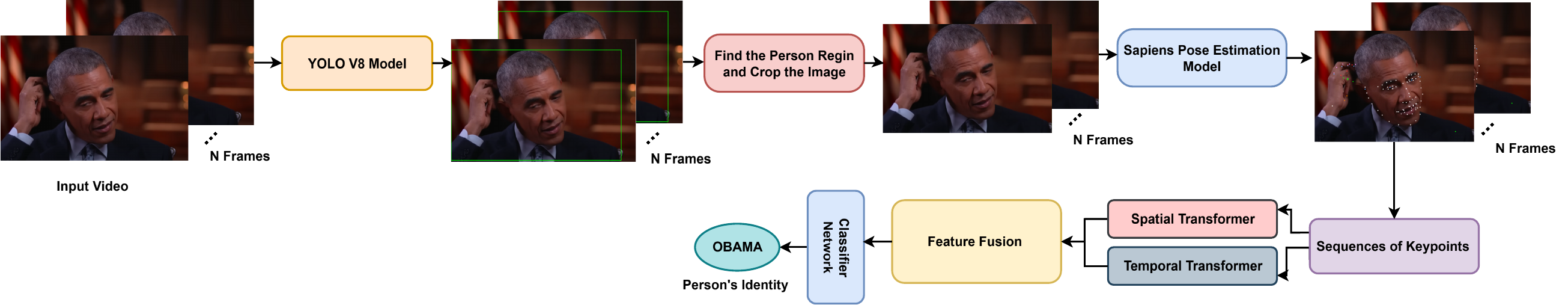}
    \caption{The block diagram of the proposed method illustrates: 1) input video; 2) person detection and localization; 3) pose estimation using the Sapiens model; 4) keypoint sequence extraction; 5) spatial and temporal transformer processing; and 6) feature fusion for transformer-based identity identification.}
    \label{fig:block_diagram}
\end{figure*}

The block diagram of the proposed method is shown in figure\ref{fig:block_diagram}. This framework consists of the following main subblocks: a person detection module for identifying and cropping the region of interest specific to the person, a pose estimation module for extracting keypoints, and spatial and temporal transformers for extracting features from the keypoints' sequences and feature fusion. In the rest of this section, we provide a detailed explanation of each component. (See Figure~\ref{fig:block_diagram}.)
\subsection{Person Detection}\label{AA}
The person detection module forms the foundation of the proposed framework. We employed the YOLOv8 model \cite{yolo}, an object detection architecture known for its accuracy, speed, and efficiency in real-time applications. Its anchor-free design and adaptive computation for variable image sizes enable reliable detection even in challenging scenarios involving occlusions or diverse lighting conditions. In our framework, YOLOv8  is utilized to isolate and extract the person’s region, ensuring high precision in identifying individuals. This step is critical for downstream tasks, as it provides clean, localized input for subsequent pose estimation and identity representation processes

\begin{figure}[h!]
    \centering
    \includegraphics[width=\linewidth]{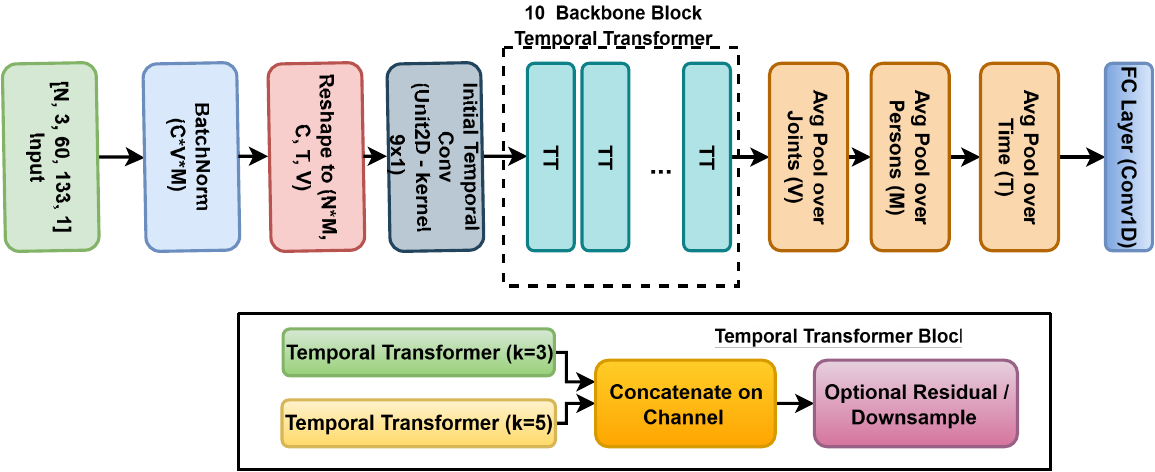}
    \caption{Multi-Scale Temporal Transformer (MS-TTR): processes inputs at multiple temporal resolutions $(k=3, k=5)$, concatenates features, and optionally applies residual connections.}
    \label{fig:ttr}
\end{figure}

\begin{figure}[h!]
    \centering
    \includegraphics[width=\linewidth]{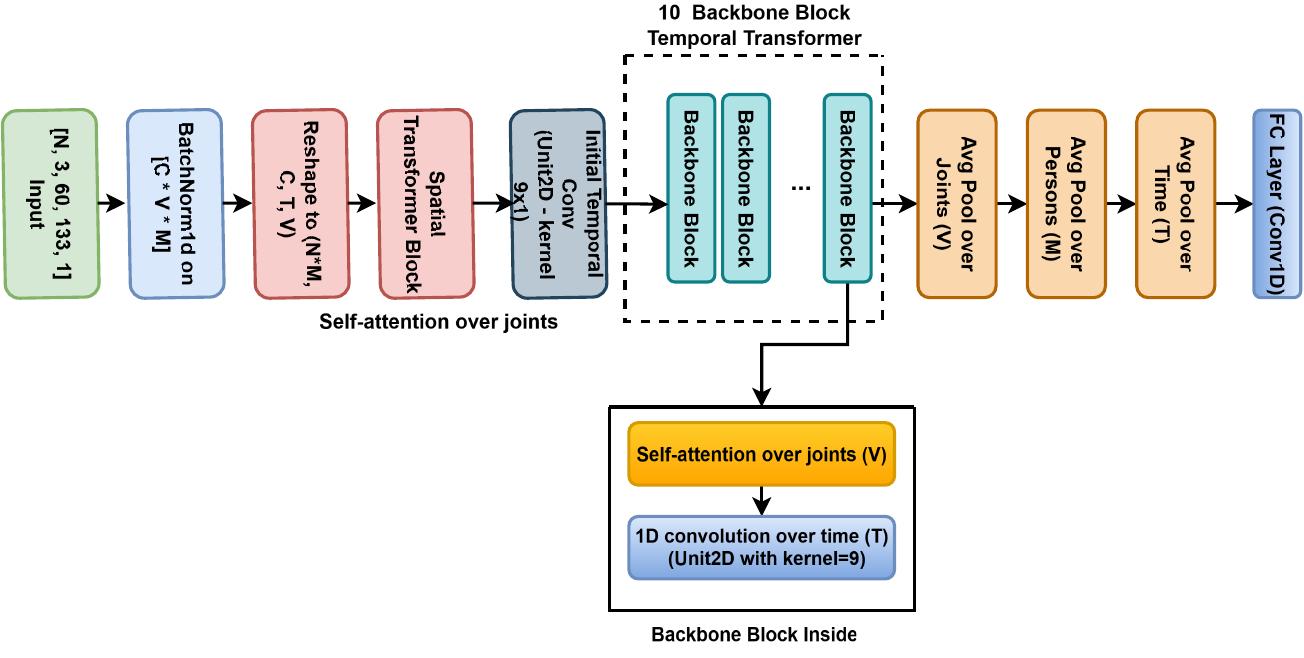}
    \caption{Spatial Transformer (STR) architecture: processes input keypoints through spatial self-attention blocks to learn flexible dependencies between body joints at each frame, followed by joint and temporal averaging before final classification.}
    \label{fig:str}
\end{figure}

\subsection{Pose Estimation}
The pose estimation module leverages the Sapiens model \cite{Sapiens}, a state-of-the-art system trained on over 300 million in-the-wild human images from the Humans-300M dataset. We utilize the Sapiens-0.3B variant (0.3 billion parameters) to extract COCO-WholeBody format keypoints, capturing detailed facial expressions, hand gestures, and body movements. The model's robust performance across varying lighting conditions and partial occlusions ensures reliable keypoint extraction in natural conversational settings. The Sapiens-0.3B model extracts 133 keypoints in COCO-WholeBody format \cite{jin2020whole}, providing comprehensive coverage of:
\begin{itemize}
\item \textbf{Facial features}: 68 points capturing detailed expressions
\item \textbf{Hand gestures}: 42 points (21 per hand) for fine-grained movements
\item \textbf{Body posture}: 17 points for overall body configuration
\item \textbf{Feet positioning}: 6 points for stance analysis
\end{itemize}

\subsection{Spatial Transformer Architecture}

The spatial transformer learns flexible, data-driven spatial dependencies between body joints at each frame, moving beyond predefined skeletal connections. Unlike graph-based methods, it constructs an attention map per layer and per sample, based on the input data.

Figure~\ref{fig:str} illustrates the spatial transformer architecture. Each frame of keypoint coordinates $X_t \in \mathbb{R}^{V \times d}$ (where $V=133$ and $d$ is the feature dimension) is transformed into \textbf{query} ($Q_t$), \textbf{key} ($K_t$), and \textbf{value} ($V_t$) representations through learned projections:

\begin{equation}
Q_t = X_t W_Q, \quad K_t = X_t W_K, \quad V_t = X_t W_V
\end{equation}

where $W_Q$, $W_K$, $W_V \in \mathbb{R}^{d \times d_k}$ are learnable weight matrices.

The \textbf{spatial self-attention} is computed by measuring pairwise joint similarities and weighting the features accordingly:
\begin{equation}
\text{SSA}(X_t) = \text{softmax}\left(\frac{Q_t K_t^T}{\sqrt{d_k}}\right) V_t
\end{equation}

This effectively aggregates information from the most informative joints for each joint in the frame. The learned attention weights highlight patterns such as face-hand coordination, postural signatures, and cross-body interactions.

\subsection{Temporal Transformer Architecture}

While the spatial transformer captures static spatial configurations, the temporal transformer focuses on the dynamics of individual joints over time. For each joint $v$, its time series $X_v \in \mathbb{R}^{T \times d}$ is embedded into queries, keys, and values:
\begin{equation}
Q_v = X_v W_Q^t, \quad K_v = X_v W_K^t, \quad V_v = X_v W_V^t
\end{equation}

The \textbf{temporal self-attention} then computes dependencies between frames:
\begin{equation}
\text{TSA}(X_v) = \text{softmax}\left(\frac{Q_v K_v^T}{\sqrt{d_k}}\right) V_v
\end{equation}

This mechanism learns which past or future frames are most relevant for the current frame, capturing patterns such as gesture onsets, pauses, rhythmic pacing, and coordination over time.

To model the hierarchical nature of conversational dynamics, we introduce a \textbf{multi-scale temporal transformer}. It processes the sequence at multiple temporal resolutions ($k=3$ and $k=5$) to simultaneously capture micro-expressions and longer gestures:
\begin{equation}
\text{MS-TTR}(X) = \text{Concat}[\text{TTR}_{k=3}(X), \text{TTR}_{k=5}(X)]
\end{equation}

where each scale subsamples the sequence as:
\begin{equation}
\text{TTR}_k(X) = \text{TSA}(X[::k, :])
\end{equation}

Figure~\ref{fig:ttr} illustrates the multi-scale design.

In both spatial and temporal branches, attention maps are learned \emph{per layer and per sample}, ensuring adaptation to individual-specific patterns. Together, these modules form a complementary two-stream framework for robust person identification from conversational keypoint dynamics.

\subsubsection{Integration of Spatial and Temporal Features}

To leverage both static postural cues and dynamic motion patterns, we employ a feature‐level fusion strategy. Let
\(\mathbf{f}_S \in \mathbb{R}^d\) and \(\mathbf{f}_T \in \mathbb{R}^d\)
be the L2‐normalized embeddings from the Spatial Transformer (STR) and Temporal Transformer (TTR), respectively. We form the concatenated feature vector
\[
\mathbf{f}_{\mathrm{fus}} \;=\; [\,\mathbf{f}_S \,\|\, \mathbf{f}_T\,]\;\in\;\mathbb{R}^{2d}.
\]

This is passed through a three‐layer fusion classifier with dropout and batch normalization:

\begin{align*}
\mathbf{h}_1 &= \mathrm{ReLU}\bigl(\mathrm{BatchNorm}(W_1\,\mathbf{f}_{\mathrm{fus}} + b_1)\bigr), \\
\mathbf{h}_1' &= \mathrm{Dropout}(\mathbf{h}_1), \\[6pt]
\mathbf{h}_2 &= \mathrm{ReLU}\bigl(\mathrm{BatchNorm}(W_2\,\mathbf{h}_1' + b_2)\bigr), \\
\mathbf{h}_2' &= \mathrm{Dropout}(\mathbf{h}_2), \\[6pt]
\hat{\mathbf{y}} &= W_3\,\mathbf{h}_2' + b_3,
\end{align*}

where \(W_1\in\mathbb{R}^{2d\times 2d}\), \(W_2\in\mathbb{R}^{2d\times d}\), \(W_3\in\mathbb{R}^{d\times C}\), and each Dropout has \(p=0.2\).  This end‐to‐end fusion classifier learns to combine spatial and temporal cues while regularizing via dropout.

We train the dual‐stream system with a combined loss:
\[
\mathcal{L}_{\mathrm{total}}
= \mathcal{L}_{\mathrm{STR}}
+ \mathcal{L}_{\mathrm{TTR}}
+ \mathcal{L}_{\mathrm{FUSION}},
\]
where each \(\mathcal{L}\) is the cross‐entropy loss of the corresponding stream or fusion output. This weighted formulation ensures balanced learning across the spatial, temporal, and fused components without yet reporting the empirical results.

\section{Experiments}

\subsection{Dataset and Setup}

We conduct experiments on a subset of the CANDOR conversational corpus \cite{candor}, which provides a large collection of spontaneous, natural interactions. From the full corpus of over 850 hours and 1,656 conversations, we select 114 unique individuals who each participated in multiple sessions. For each individual, we use their longest recording session, segmented into 3--4 second utterances based on transcription boundaries. This setup allows us to focus on keypoint dynamics within short, natural segments while controlling for potential overlap.  

We divide the utterances with an 80:20 train–test split at the utterance level, ensuring no overlap of utterances across splits. For computational efficiency, we sample every second frame, yielding 30-frame sequences at 30fps. Pose keypoints in COCO WholeBody format (133 points per frame) are extracted with the Sapiens-0.3B model. This keypoint-only representation removes static appearance cues and isolates dynamic behavioral patterns.  

\subsection{Implementation Details}

Training is performed on an NVIDIA RTX 4080 GPU. Models are optimized with Adam for 120 epochs, starting from a learning rate of 0.001. Batch normalization and dropout (0.2) are used for regularization. We compare models trained from scratch (domain-specific) and models initialized with weights pre-trained for action recognition from videos \cite{Plizzari}. Pre-trained weights for STR and TTR are loaded from the publicly available implementation in \cite{Plizzari}.

We experiment with several transformer-based architectures:
\begin{itemize}
    \item \textbf{Spatial Transformer (STR):} processes each frame independently with a self-attention mechanism over joints to learn postural configurations.
    \item \textbf{Temporal Transformer (TTR):} models motion dynamics over time at a single temporal resolution ($k=9$).
    \item \textbf{Multi-Scale Temporal Transformer (MS-TTR):} processes sequences at two resolutions ($k=3$ and $k=5$) to capture both micro- and macro-gestures.
    \item \textbf{STR + TTR Fusion:} feature-level fusion of spatial and temporal embeddings, followed by a joint classifier.
\end{itemize}
\begin{table}[htbp]
\caption{Performance comparison of spatial, temporal, and fused transformer architectures for person identification.}
\begin{center}
\begin{tabularx}{\columnwidth}{|X|c|c|}
\hline
\textbf{Method} & \textbf{Accuracy (\%)} & \textbf{Training Strategy} \\
\hline
\hline
\multicolumn{3}{|c|}{\textbf{Spatial Transformer}} \\
\hline
STR (Pretrained) & 80.12 & Transfer Learning \\
STR (From Scratch) & 95.74 & Domain-Specific \\
\hline
\hline
\multicolumn{3}{|c|}{\textbf{Temporal Transformer}} \\
\hline
TTR (Pretrained) & 63.61 & Transfer Learning \\
TTR (From Scratch) & 82.22 & Domain-Specific $k=9$ \\
Multi-Scale TTR   & 93.90 & Multi-Scale $k=3, 5$ \\
\hline
\hline
\multicolumn{3}{|c|}{\textbf{Dual-Stream Fusion}} \\
\hline
STR + TTR (Feature Fusion) & \textbf{98.03} & Feature-Level Fusion \\
\hline
\end{tabularx}
\label{tab:spatial_temporal}
\end{center}
\end{table}
For TTR, we also evaluate the effect of explicitly adding velocity features, computed as:
\begin{equation}
\text{velocity} = \text{joint\_coords}[:,:,1:] - \text{joint\_coords}[:,:,:-1]
\end{equation}

which emphasizes frame-to-frame positional changes \cite{Vuillecard}.

\begin{table}[htbp]
\caption{Impact of Velocity Features on Temporal Transformer}
\begin{center}
\begin{tabularx}{\columnwidth}{|X|c|c|}
\hline
\textbf{Method} & \textbf{Accuracy (\%)} & \textbf{Change} \\
\hline
TTR (k=9) (From Scratch) & 82.22 & Baseline \\
TTR (k=9) + Velocity & 75.04 & -7.18\% \\
\hline
\end{tabularx}
\label{tab:velocity_analysis}
\end{center}
\end{table}

\begin{table}[htbp]
\caption{Computational Efficiency Analysis}
\begin{center}
\begin{tabularx}{\columnwidth}{|X|c|c|c|}
\hline
\textbf{Model} & \textbf{Params (M)} & \textbf{FLOPs (G)} & \textbf{FPS} \\
\hline
Spatial Transformer & 3.29 & 6.908 & 165.64 \\
Temporal Transformer (k=9) & 1.89 & 3.374 & 52.93 \\
Multi-Scale Temporal Transformer (k=3, k=5) & 1.837 & 4.894 & 64.46 \\
STR + TTR (Feature Fusion)& 6.069 & 11.802 & 46.74 \\
\hline
\end{tabularx}
\label{tab:efficiency}
\end{center}
\end{table}
\begin{table*}[htbp]
\caption{Comparison of Methods for Person Identification}
\centering
\begin{tabularx}{\textwidth}{|l|X|c|X|X|X|}
\hline
\textbf{Ref.} & \textbf{Dataset} & \textbf{Number of Identities} & \textbf{Features} & \textbf{Classifier} & \textbf{Accuracy} \\
\hline
\cite{Kay} & CASME II, SMIC, SAMM & 26, 16, 32 & SlowFast CNN & Fully connected layer & 94.95\%, 89.61\%, 87.4\% \\
\hline
\cite{Saracbasi} & MYFED & 50 & Statistical facial dynamics features & KNN, LSTM & KNN: 88.1\%, LSTM: 87.8\% \\
\hline
\cite{Papadopoulos} & BU4DFE & 101 & Spatio-temporal graph features & Spatio-Temporal Graph Convolutional Network (ST-GCN) & 88.45\% \\
\hline
\cite{Haamer} & Self-collected & 61 & VGG-face CNN and geometric features & LSTM & 96.2\% \\
\hline
Ours & Candor’s subset & 114 & extracted features from 133 keypoints using TTR and STR &  Three-layer fusion classifier & 98.03\% \\
\hline
\end{tabularx}
\label{tab:comparison}
\end{table*}

\subsection{Experimental Results}
Table~\ref{tab:spatial_temporal} summarizes the results. Domain-specific training consistently outperforms pre-trained weights across both STR and TTR, with particularly dramatic improvements for TTR (82.22\% vs 63.61\%). This performance gap likely stems from the fundamental mismatch between the pre-training task—action recognition emphasizing large, categorical movements—and our target domain of subtle conversational dynamics characterized by micro-expressions and nuanced gestural patterns.

The STR achieves exceptional single-stream accuracy (95.74\%), demonstrating that spatial joint configurations within individual frames encode highly discriminative identity signatures. These postural patterns—including habitual head tilts, shoulder positioning, and hand-face spatial relationships—appear remarkably stable within conversational contexts.

The temporal modeling results reveal important insights about motion sampling strategies. The standard TTR with k=9 subsampling achieves 82.22\% accuracy despite processing only 3-4 frames from each 30-frame sequence. This sparse sampling captures only coarse postural transitions, missing the rich temporal dynamics of natural conversation. The multi-scale variant (k=3, k=5) dramatically improves performance to 93.90\% by processing motion at complementary temporal resolutions: k=3 preserves medium-scale movements like hand gestures and facial expressions (sampling ~10 frames), while k=5 captures slightly coarser gestural arcs (~6 frames). This 11.68\% improvement demonstrates the critical importance of multi-resolution temporal modeling for capturing both micro-expressions and extended gestural phrases.

Interestingly, incorporating explicit velocity features degrades performance (75.04\%), suggesting that conversational identity manifests not only through motion but also through characteristic pauses, held postures, and timing patterns that velocity-based representations inherently suppress.

The feature-level fusion achieves 98.03\% accuracy, confirming that spatial configurations and temporal dynamics encode complementary identity information that, when combined, enable near-perfect recognition even from brief conversational segments.

\subsection{Computational Analysis}
Table~\ref{tab:efficiency} presents the computational requirements of each architecture. The STR processes individual frames independently, achieving the highest throughput (165.64 FPS) suitable for real-time applications. The multi-scale TTR demonstrates remarkable efficiency, requiring fewer parameters (1.837M) than the single-scale variant (1.89M) while delivering superior accuracy. This parameter efficiency likely results from shared backbone layers between scales and optimized feature concatenation strategies.

The complete dual-stream system, while computationally intensive (6.069M parameters, 46.74 FPS), maintains real-time viability for practical deployment. The trade-off between computational cost and accuracy is favorable, with the 2.29\% accuracy gain from fusion justifying the increased complexity for security-critical applications.

\subsection{Comparative Discussion}
Table~\ref{tab:comparison} contextualizes our results against existing dynamic recognition methods. Direct comparison remains challenging due to fundamental differences in experimental settings—prior works predominantly utilize controlled laboratory environments with posed emotions, smaller subject pools (typically 16-101 identities), and often incorporate appearance features alongside dynamics.

Despite these differences, our method achieves state-of-the-art performance (98.03\%) while addressing a substantially more challenging scenario: 114 identities in spontaneous conversation using only skeletal keypoints. This represents a significant advance in both scale and naturalism. Notably, we surpass the previous best result on naturalistic data [24] by 1.83\% while nearly doubling the number of identities and eliminating all appearance cues.

\section{Discussion}
Our results demonstrate that conversational behavior contains robust identity signatures that can be extracted through transformer architectures operating on skeletal keypoints alone. The exceptional spatial transformer performance (95.74\%) reveals that individuals maintain highly consistent postural configurations during conversation—including characteristic head-hand alignments and shoulder positions—that serve as reliable identity markers even when facial features are unavailable.

The temporal analysis illuminates the multi-scale nature of conversational dynamics. Single-scale sampling (k=9) achieves limited success (82.22\%) by capturing only coarse postural transitions across 3-4 frames. In contrast, the multi-scale approach (93.90\%) successfully models the full spectrum of conversational movement: rapid micro-expressions (3-5 frames), gestural arcs (10-15 frames), and extended postural shifts. Notably, velocity features degrade performance (75.04\%), indicating that conversational identity emerges not just from movement but from the interplay of motion, stillness, and timing—the characteristic pause before speaking or duration of a smile may be as identifying as the gestures themselves.

The near-perfect fusion accuracy (98.03\%) confirms that spatial and temporal features encode complementary identity aspects: spatial configurations capture postural habits while temporal patterns reveal behavioral dynamics and rhythm. This finding has immediate applications for privacy-preserving identification, authentication in challenging visual conditions, and behavioral analysis systems that must operate without relying on facial appearance.

\section{Limitations and Future Work}
Our study is constrained by several factors that suggest directions for future research: (1) the dataset includes only 114 speakers from single sessions, limiting generalization assessment; (2) cultural variations in conversational behavior remain unexplored; (3) the 3-4 second utterance segments may miss longer-term behavioral patterns; and (4) alternative motion representations beyond raw velocity could better capture conversational dynamics. Future work should address these limitations through larger cross-cultural datasets, extended temporal modeling, and integration with complementary biometric modalities.

\section{Conclusion}
We presented a transformer-based framework for person identification from conversational dynamics using only skeletal keypoints. Our key contribution is demonstrating that natural conversational behavior—without any posed expressions or appearance features—contains sufficient identity information for highly accurate recognition. The spatial transformer captures stable postural configurations (95.74\%), while the multi-scale temporal transformer models motion dynamics across multiple timescales (93.90\%). Their fusion achieves 98.03\% accuracy on 114 speakers, establishing conversational dynamics as a viable behavioral biometric.

By isolating behavioral patterns from appearance, this work advances our understanding of identity-specific movement in natural interaction and provides a foundation for privacy-preserving identification systems. The success of keypoint-only recognition suggests that how we move during conversation is as distinctive as traditional biometric features, opening new avenues for behavioral analysis and human-computer interaction.


\begin{thebibliography}{00}
\bibitem{baisa} N. L. Baisa, ``Joint person identity, gender and age estimation from hand images using deep multi-task representation learning,'' in \textit{Proc. 12th Int. Workshop Biometrics and Forensics (IWBF)}, 2024, pp. 01--06.

\bibitem{gait} V. Rani and M. Kumar, ``Human gait recognition: A systematic review,'' \textit{Multimedia Tools and Applications}, vol. 82, no. 24, pp. 37003--37037, 2023.


\bibitem{voice} H. Kheddar, M. Hemis, and Y. Himeur, ``Automatic speech recognition using advanced deep learning approaches: A survey,'' \emph{Information Fusion}, vol. 109, p. 102422, 2024.

\bibitem{Alrawili} R. Alrawili, A. A. S. AlQahtani, and M. K. Khan, ``Comprehensive survey: Biometric user authentication application, evaluation, and discussion,'' \textit{Computers and Electrical Engineering}, vol. 119, p. 109485, 2024.

\bibitem{Battisti} A. Battisti, E. van den Bold, A. Göhring, F. Holzknecht, and S. Ebling, ``Person identification from pose estimates in sign language,'' University of Zurich, 2024.

\bibitem{L. Wang} L. Wang, H. Ning, T. Tan, and W. Hu, ``Fusion of static and dynamic body biometrics for gait recognition,'' \textit{IEEE Trans. Circuits Syst. Video Technol.}, vol. 14, no. 2, pp. 149--158, 2004.
\bibitem{Farhadipour} A. Farhadipour, M. Chapariniya, T. Vukovic, and V. Dellwo, ``Comparative analysis of modality fusion approaches for audio-visual person identification and verification,'' in \textit{Proc. 7th Int. Conf. Natural Language and Speech Processing (ICNLSP)}, 2024, pp. 168--177.

\bibitem{Hill} H. Hill and A. Johnston, ``Categorizing sex and identity from the biological motion of faces,'' Current Biol., vol. 11, no. 11, pp. 880--885, 2001.

\bibitem{Girges} C. Girges, J. Spencer, and J. O'Brien, ``Categorizing identity from facial motion,'' Quarterly Journal of Experimental Psychology, vol. 68, no. 9, pp. 1832--1843, 2015.

\bibitem{Dobs} K. Dobs, I. Bülthoff, and J. Schultz, ``Identity information content depends on the type of facial movement,'' \textit{Scientific Reports}, vol. 6, no. 1, pp. 34301, 2016.

\bibitem{Papadopoulos} K. Papadopoulos, A. Kacem, D. Aouada \textit{et al.}, ``Face-GCN: A graph convolutional network for 3D dynamic face recognition,'' in \textit{Proc. 8th Int. Conf. Virtual Reality (ICVR)}, 2022, pp. 454--458.

\bibitem{X. Zhang} X. Zhang, L. Yin, J. F. Cohn, S. Canavan, M. Reale, A. Horowitz, and P. Liu, ``A high-resolution spontaneous 3D dynamic facial expression database,'' in \textit{Proc. 10th IEEE Int. Conf. and Workshops on Automatic Face and Gesture Recognition (FG)}, Shanghai, China, 2013, pp. 1--6.


\bibitem{Kay} T. Kay, Y. Ringel, K. Cohen, M.-A. Azulay, and D. Mendlovic, ``Person recognition using facial micro-expressions with deep learning,'' arXiv preprint arXiv:2306.13907, 2023.

\bibitem{CASME} W.-J. Yan, X. Li, S.-J. Wang, G. Zhao, Y.-J. Liu, Y.-H. Chen, and X. Fu, ``CASME II: An improved spontaneous micro-expression database and the baseline evaluation,'' PLoS One, vol. 9, no. 1, p. e86041, 2014.


\bibitem{Samm} A. K. Davison, C. Lansley, N. Costen, K. Tan, and M. H. Yap, ``Samm: A spontaneous micro-facial movement dataset,'' \textit{IEEE Transactions on Affective Computing}, vol. 9, no. 1, pp. 116--129, 2016.

\bibitem{Saracbasi} Z. N. Saracbasi, C. E. Erdem, M. Taskiran, and N. Kahraman, ``MYFED: a dataset of affective face videos for investigation of emotional facial dynamics as a soft biometric for person identification,'' \textit{Machine Vision and Applications}, vol. 36, no. 1, p. 8, 2025.
\bibitem{rack2022comparison} C. Rack, A. Hotho, and M. E. Latoschik, ``Comparison of data encodings and machine learning architectures for user identification on arbitrary motion sequences,'' in \textit{Proc. IEEE Int. Conf. Artificial Intelligence and Virtual Reality (AIVR)}, 2022, pp. 11--19.

\bibitem{yolo} G. Jocher, A. Chaurasia, and J. Qiu, ``Ultralytics YOLOv8,'' version 8.0.0, 2023. [Online]. Available: https://github.com/ultralytics/ultralytics

\bibitem{Sapiens} R. Khirodkar, T. Bagautdinov, J. Martinez, S. Zhaoen, A. James, P. Selednik, S. Anderson, and S. Saito, ``Sapiens: Foundation for human vision models,'' in \textit{Proc. European Conf. Comput. Vision (ECCV)}, pp. 206--228, Springer, 2024.

\bibitem{jin2020whole} S. Jin, L. Xu, J. Xu, C. Wang, W. Liu, C. Qian, W. Ouyang, and P. Luo, ``Whole-body human pose estimation in the wild,'' in \textit{Proc. Eur. Conf. Comput. Vis. (ECCV)}, 2020, pp. 196--214.

\bibitem{candor} A. Reece, G. Cooney, P. Bull, C. Chung, B. Dawson, C. Fitzpatrick, T. Glazer, D. Knox, A. Liebscher, and S. Marin, ``The CANDOR corpus: Insights from a large multimodal dataset of naturalistic conversation,'' \textit{Science Advances}, vol. 9, no. 13, p. eadf3197, 2023.

\bibitem{Plizzari} C. Plizzari, M. Cannici, and M. Matteucci, ``Skeleton-based action recognition via spatial and temporal transformer networks,'' Computer Vision and Image Understanding, vol. 208, p. 103219, 2021.

\bibitem{Vuillecard} P. Vuillecard, A. Farkhondeh, M. Villamizar, and J.-M. Odobez, ``Ccdb-hg: Novel annotations and gaze-aware representations for head gesture recognition,'' in \textit{Proc. 2024 IEEE 18th Int. Conf. Automatic Face and Gesture Recognition (FG)}, pp. 1--9, 2024.

\bibitem{Haamer} R. E. Haamer, K. Kulkarni, N. Imanpour, M. A. Haque, E. Avots, M. Breisch, K. Nasrollahi, S. Escalera, C. Ozcinar, X. Baro, \textit{et al.}, ``Changes in facial expression as biometric: a database and benchmarks of identification,'' in \textit{Proc. 13th IEEE Int. Conf. Automatic Face \& Gesture Recognition (FG)}, Xi'an, China, 2018, pp. 621--628.








\end{thebibliography}
\end{document}